\documentclass[10pt,twocolumn,letterpaper]{article}

\usepackage[pagenumbers]{cvpr} 

\usepackage{booktabs}        
\usepackage{xcolor} 
\usepackage{multicol} 
\usepackage{multirow} 
\usepackage{amsmath} 
\usepackage{graphicx} 
\usepackage{bbding}   
\usepackage{makecell}   
\usepackage{diagbox} 
\usepackage{colortbl} 
\usepackage{amssymb}

\makeatletter
\renewcommand\paragraph{\@startsection{paragraph}{4}{\z@}%
                                     {.8ex \@plus .2ex \@minus .1ex}%
                                     {-1em}%
                                     {\normalfont\normalsize\bfseries}}
\makeatother

\definecolor{cvprblue}{rgb}{0.21,0.49,0.74}
\usepackage[pagebackref,breaklinks,colorlinks,allcolors=cvprblue]{hyperref}

\def\paperID{} 
\def\confName{CVPR}
\def\confYear{2026}

\title{MFEN: Multi-Frequency Expert Network for Visible-Infrared Person Re-ID}

\author{
Xulin Li\textsuperscript{1,2},
Yan Lu\textsuperscript{3},
Bin Liu\textsuperscript{1,2}\textsuperscript{*},
Qinhong Yang\textsuperscript{1,2},
Qi Chu\textsuperscript{1,2},
Tao Gong\textsuperscript{1,2},
Nenghai Yu\textsuperscript{1,2}\\
\textsuperscript{1}University of Science and Technology of China, China\\
\textsuperscript{2}Anhui Province Key Laboratory of Digital Security, China\\
\textsuperscript{3}The Chinese University of Hong Kong, China\\
{\tt\small lxlkw@mail.ustc.edu.cn, yanlu@cuhk.edu.hk, flowice@ustc.edu.cn}\\
{\tt\small qhyang233@mail.ustc.edu.cn, \{qchu,tgong,ynh\}@ustc.edu.cn}\\
{\small \textsuperscript{*}Corresponding author}
}

\begin{document}
\maketitle
\begin{abstract}
Visible-infrared person re-identification (VI-ReID) is challenging due to the large modality discrepancy between visible and infrared images.
We contend that this discrepancy is largely related to differing lighting conditions, including differences in light wavelength and light source type.
Recently, frequency-based VI-ReID approaches have achieved notable success because frequency information can better extract identity-relevant contours and details while excluding irrelevant lighting and color.
However, existing methods either do not distinguish different frequency bands or focus on only one band, which is insufficient under diverse lighting conditions.
To perform comprehensive frequency-domain learning, we propose a Multi-Frequency Expert Network (MFEN) that enables multi-frequency modulation and adaptively combines different bands through a mixture-of-experts design.
We further introduce Random Frequency Augmentation (RFA) and Frequency Auxiliary Optimization (FAO) to better train MFEN.
The three modules are complementary and jointly capture critical frequency-domain details for robust representation learning.
Extensive experiments on three VI-ReID datasets demonstrate the effectiveness of our approach.
\end{abstract}
 
\section{Introduction}
\label{sec:intro}

\begin{figure}[!t]
    \centering
    \includegraphics[width=1.0\linewidth]{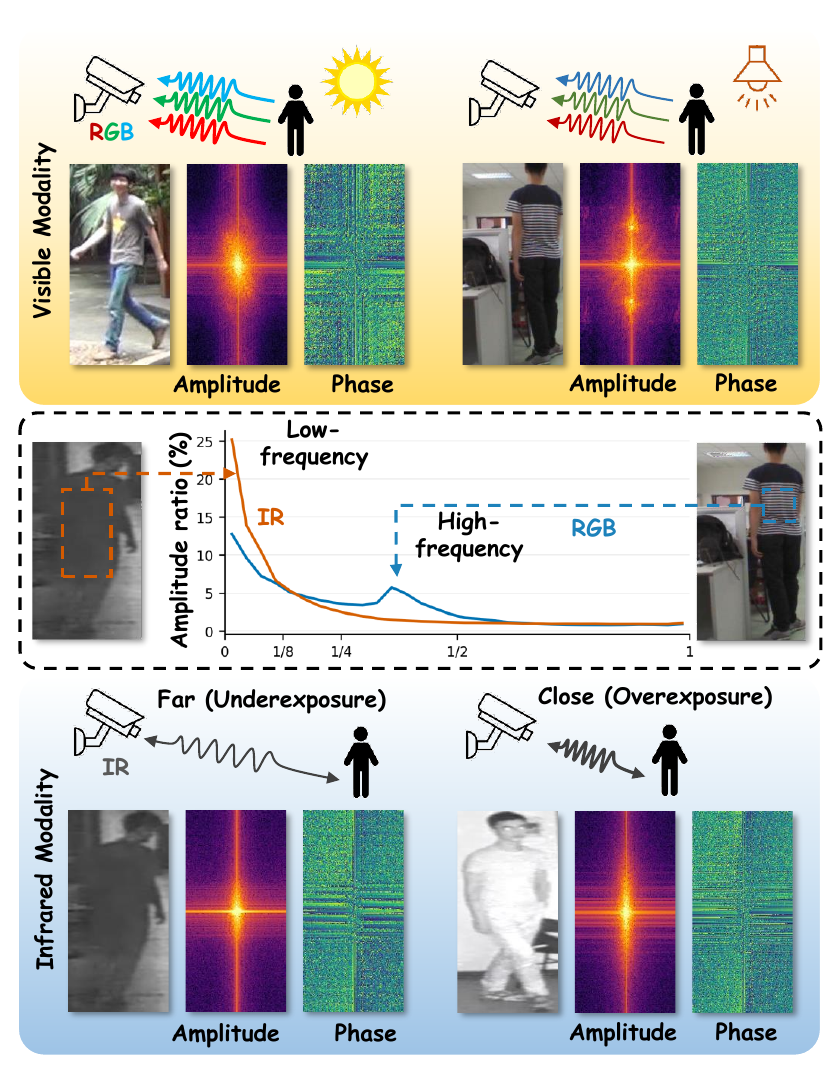}
    \caption{
    Different lighting conditions lead to gaps in visible and infrared modality data, including color differences caused by light wavelength and brightness differences resulting from the type of light source.
    The frequency domain can better distinguish identity cues from irrelevant lighting and color.
    }
    \label{fig:intro}
    \vspace{0mm}
\end{figure}

Person re-identification (ReID)~\cite{ye2021deep,li2023clip,zheng2024versatile,li2025ATreid} aims to match images of the same individual captured by different cameras and is widely used in surveillance systems.
To enable 24-hour ReID, infrared (IR) cameras are used to capture pedestrians at night, motivating visible-infrared person re-identification (VI-ReID)~\cite{wu2017rgb,zhang2023diverse}.
Different from single-modality ReID, VI-ReID must address both the RGB-IR modality gap and the large intra-modality variation of IR images.
Different lighting conditions are a major source of the modality gap between RGB and IR images. 
For example, RGB images capture light across red, green, and blue bands, providing rich color information, whereas IR images capture only one band~\cite{ye2021channel,zhao2021joint,liu2021sfanet}. 

Beyond wavelength differences, we argue that the discrepancy of light sources is also important but underexplored.
As shown in \cref{fig:intro}, RGB images are captured under diverse light sources, including sunlight, artificial light, and ambient reflected light, yielding clearer details.
In contrast, IR images captured by surveillance cameras rely solely on the camera's light source, which often results in underexposed and overexposed images.
Consequently, the variations in brightness caused by differing light sources significantly exacerbate the intra-identity gap, posing a critical challenge for VI-ReID.

Fortunately, frequency-domain representations explicitly capture illumination-related image variations.
The amplitude spectrum primarily encodes critical cues related to lighting and color~\cite{ma2019infrared}. Underexposed and overexposed areas are mainly reflected in low-frequency components, whereas mid- and high-frequency components typically correspond to image details and edges~\cite{zhang2023pha,li2024adaptive}.
Therefore, frequency information is promising for mitigating illumination-induced modality discrepancy and has shown success in VI-ReID.
However, identity-related and illumination-related cues are distributed across multiple bands under complex lighting conditions. Severely overexposed samples require suppressing dominant low-frequency illumination, while blurred or low-contrast samples rely more on informative mid- and high-frequency details. Thus, the optimal band is sample-dependent. Existing frequency-based methods~\cite{zhang2023protohpe,zhang2024frequency,li2024frequency,chen2024visible} either modulate the entire frequency range or focus exclusively on fixed high-frequency regions, which is insufficient for such adaptive selection.

To address this gap, we propose a novel Multi-Frequency Expert Network (MFEN), which extracts identity-relevant contours and details while suppressing irrelevant lighting and color. 
Specifically, our MFEN constructs multiple frequency modulation experts, each corresponding to a different frequency band. Information from these experts is dynamically blended by a gating function, enabling each sample to adaptively select the most relevant frequency cues instead of relying on a fixed-band prior.
To effectively train the MFEN, we further introduce a Random Frequency Augmentation (RFA) and a Frequency Auxiliary Optimization (FAO). 
Our RFA technique exchanges the low-frequency components of the amplitude spectrum between two modal images to reduce lighting and color discrepancies while preserving structural details and avoiding obvious artifacts.
Our FAO trains the model in both the spatial and frequency domains.
The key idea of FAO lies in leveraging frequency-domain statistics to complement the final spatial features, while applying mirrored loss constraints on both representations to further facilitate cross-modal feature learning.
These modules are complementary and jointly capture critical frequency-domain details for robust VI-ReID representations.

Our main contributions can be summarized as follows:

\noindent $\bullet$ 
We investigate the key challenges of VI-ReID and identify lighting differences as a significant factor.
To effectively fuse frequency information, we propose a novel Multi-Frequency Expert Network (MFEN), which adaptively captures critical frequency-domain details from multiple bands on a per-sample basis and complements the spatial domain.

\noindent $\bullet$
To mitigate cross-modal discrepancy at the data level, we propose Random Frequency Augmentation (RFA) to simulate the lighting patterns of another modality.
To further reduce the modality gap, we introduce Frequency Auxiliary Optimization (FAO) to incorporate frequency-domain constraints during optimization.

\noindent $\bullet$
The three proposed modules are complementary.
Extensive experiments on three large-scale VI-ReID datasets demonstrate the superiority of our method against the state-of-the-art.

\section{Related Work}   
\label{sec:relate}

\paragraph{Visible-Infrared Person Re-identification.}
Compared to standard single-modality person re-identification, VI-ReID must address a significant modality gap.
Early works~\cite{mao2017aligngan,dai2018cross,wang2020cross} bridged this gap by transforming one modality to the other, but generated images often struggle to preserve accurate identity information due to the lack of pose-aligned pairs.
Other methods~\cite{tan2024rle,cui2024dma,ye2021channel,qian2025visible,zhao2021joint} reduce modality differences through data augmentation. For example, Ye \etal~\cite{ye2021channel} generate color-irrelevant images by randomly exchanging color channels, and Qian \etal~\cite{qian2025visible} stitch cross-modal patches to form new samples.
Feature-space methods~\cite{feng2023shape,kim2023partmix,wu2021discover,zhang2023diverse} instead rely on carefully designed enhancement modules and cross-modality losses. Wu \etal~\cite{wu2021discover} introduced modality alleviation and pattern alignment modules, while Zhang \etal~\cite{zhang2023diverse} utilized diverse embeddings and multi-stage feature aggregation.
Unlike these methods, our RFA, MFEN, and FAO explicitly exploit frequency-domain cues to mitigate lighting-induced modality discrepancy.

\paragraph{Frequency Domain Learning.}
Frequency analysis has long been important in digital image processing, and frequency-domain learning has recently become prevalent in computer vision~\cite{zhong2022detecting,yao2022wave,miao2023f,chen2025frequency}.
%
%
%
%
%

Recently, frequency-domain learning has also attracted increasing attention in person re-identification~\cite{li2024adaptive,li2024frequency,zhang2024frequency,zhang2023pha,chen2024visible,zhang2023protohpe}. Existing methods either model the whole spectrum or emphasize fixed high-frequency components~\cite{li2024adaptive,li2024frequency,zhang2024frequency,zhang2023pha,chen2024visible,zhang2023protohpe}.
Chen \etal~\cite{chen2024visible} facilitated multi-level feature fusion in both spatial and frequency domains. Methods~\cite{li2024frequency,zhang2024frequency} explored modality-invariant learning by amplitude-phase decomposition, while others~\cite{li2024adaptive,zhang2023pha,zhang2023protohpe} focused on exploiting key high-frequency components for identification.
In contrast, MFEN performs sample-wise frequency selection, enabling more flexible handling of frequency variations under different lighting conditions.
More specifically, prior VI-ReID methods such as FDMNet~\cite{li2024frequency}, FDNM~\cite{zhang2024frequency}, and DSSF$^3$~\cite{chen2024visible} mainly rely on unified frequency-domain transformation or fusion, whereas MFEN explicitly decomposes the feature map into disjoint bands and performs sample-wise adaptive fusion through gating. Our design also differs from FDConv~\cite{chen2025frequency}, which decomposes convolution kernels for dense prediction. MFEN instead decomposes spatial feature maps, directly aligning the modulation with the cross-modal spatial frequency discrepancy in VI-ReID.

%

\begin{figure*}[!t]
    \centering
    \includegraphics[width=0.99\linewidth]{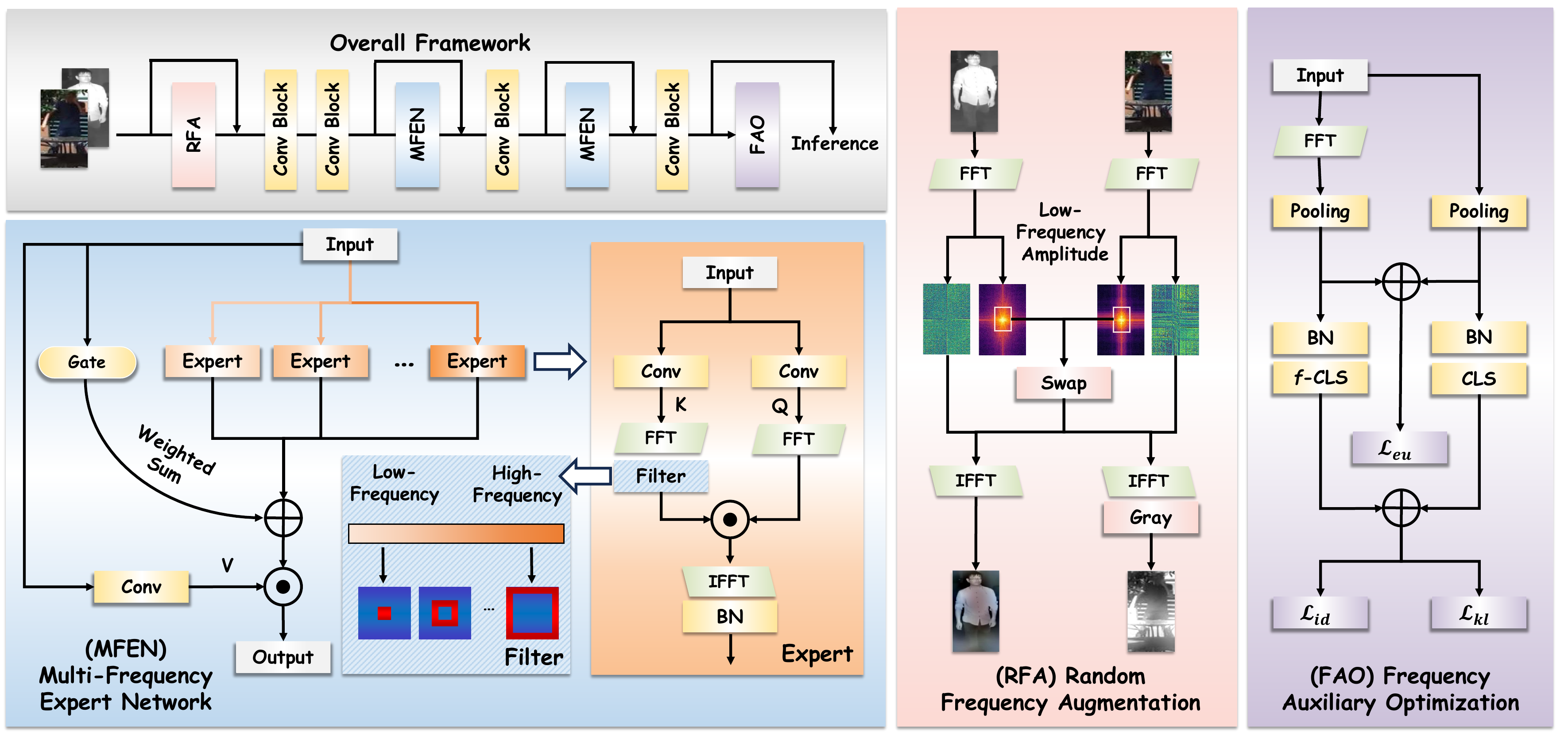}
    \caption{
    The pipeline of the proposed Multi-Frequency Expert Network (MFEN).
    Three modules are complementary to each other and train the model in the frequency domain to achieve robust and discriminative representations.
    FFT and IFFT denote the Fast Fourier Transform and its inverse operator, respectively.
    }
    \label{fig:method}
    \vspace{0mm}
\end{figure*}

\section{Method}  
\label{sec:method}

\paragraph{Overview.}
The pipeline of our method is shown in \cref{fig:method}.
Visible and infrared images are first input into the Random Frequency Augmentation (RFA) module to reduce the illumination discrepancy between the two modalities at the image level. 
Subsequently, cues from the frequency domain are extracted and fused through the model-level Multi-Frequency Expert Network (MFEN) module. 
Finally, the Frequency Auxiliary Optimization (FAO) module performs frequency domain cross-modal feature learning at the optimization level. 
These modules jointly facilitate robust cross-modality feature learning.
During testing, the augmentation and auxiliary losses are removed, and the learned network extracts the final descriptor from the spatial feature and its complementary frequency statistic.

\subsection{Random Frequency Augmentation}

Visible images capture three wavelengths and form RGB images, whereas infrared (IR) images contain only one channel.
This image-level disparity increases the burden on the feature extractor.
Simple grayscale transforms can effectively minimize the visual discrepancies between RGB and infrared (IR) images from a channel (color) perspective.
Some data augmentation techniques~\cite{ye2021channel} further refine this strategy through random channel selection.
We argue that frequent underexposure and overexposure in IR images still create a significant gap from grayscale RGB images.

Therefore, we propose the Random Frequency Augmentation (RFA) module, which transforms images to simulate the lighting patterns of another modality, including variations in both color and brightness.
Since color and brightness are primarily encoded in the amplitude component in the Fourier domain~\cite{zhang2024dmfourllie}, we swap the amplitude component of RGB and IR images in the training batch to reduce the cross-modal illumination gap.
However, the full amplitude spectrum also contains high-frequency structural energy that should remain aligned with the original phase.
Therefore, swapping the entire amplitude can cause severe artifacts and distort identity structures.
We instead swap only the low-frequency amplitude, which mainly controls global illumination and tone, while preserving the high-frequency amplitude and the original phase.

\paragraph{Image to Frequency Domain.}
Given an image $x \in \mathbb{R}^{H\times W \times 3}$, we adopt the discrete Fourier transform $\mathcal{F}(\cdot)$ on the spatial dimension of each channel to convert the image to Fourier space. 
The obtained frequency representations $\mathcal{F}(x) \in \mathbb{R}^{H\times W \times 3}$ are expressed as:

\begin{equation}   
\begin{aligned} 
    \mathcal{F}(x)(u,v) &= \sum_{h=0}^{H-1}\sum_{w=0}^{W-1}x(h,w) e^{-j2\pi (\frac{uh}{H}+\frac{vw}{W})}\\
    &= \mathcal{R}(x)(u, v) + j\cdot\mathcal{I}(x)(u,v),
\end{aligned}
\label{eq:aug_fft}
\end{equation}
where $j$ represents the imaginary unit, and $\mathcal{R}(x)(u,v)$ and $\mathcal{I}(x)(u,v)$ denote the real and imaginary parts of $\mathcal{F}(x)(u,v)$, respectively.
The amplitude component $\mathcal{A}(x)(u,v)$ and phase component $\mathcal{P}(x)(u,v)$ are expressed as:
\begin{equation}   
\begin{aligned} 
    \mathcal{A}(x)(u,v) &= \sqrt{\mathcal{R}^2(x)(u,v)+\mathcal{I}^2(x)(u,v)}\\ 
    \mathcal{P}(x)(u,v) &= arctan[\frac{\mathcal{I}(x)(u,v)}{\mathcal{R}(x)(u,v)}].
\end{aligned}
\label{eq:aug_ap}
\end{equation}
The amplitude component $\mathcal{A}(x)(u,v)$ quantifies the magnitude of the frequency $(u,v)$, which preserves the lightness and color information.

\paragraph{Swap Low-Frequency Amplitude.}
Following~\cite{li2024adaptive}, after obtaining $\mathcal{A}(x)(u,v)$, we apply Gaussian low-pass and high-pass filters to obtain the low-frequency amplitude $\mathcal{A}_l(x)(u,v)$ and the high-frequency amplitude $\mathcal{A}_h(x)(u,v)$. The swap process can then be formulated as:
\begin{equation}   
\begin{aligned} 
    \mathcal{A}_s(x)(u,v) &= \mathcal{A}_l(x')(u,v) + \mathcal{A}_h(x)(u,v),
\end{aligned}
\label{eq:aug_swapa}
\end{equation}
where $x$ is the original image, $x'$ is a randomly selected image from the other modality, and $\mathcal{A}_s(x)$ represents the amplitude after swapping. 
%

\paragraph{Frequency Domain to Image.}
After obtaining $\mathcal{A}_s(x)$, we preserve the original phase to obtain the new frequency domain representation $\mathcal{F}_s(x)$:
\begin{equation}   
\begin{aligned} 
    \mathcal{F}_s(x) &= \mathcal{A}_s(x)\cdot \exp(j\cdot\mathcal{P}(x)).
\end{aligned}
\label{eq:aug_swapf}
\end{equation}
Then, we transform $\mathcal{F}_s(x)$ from the frequency domain to the spatial domain using the inverse Fourier transform $\mathcal{F}^{-1}(\cdot)$ to obtain the enhanced image $x^{aug}$, which can be formulated as:
\begin{equation}   
\begin{aligned} 
    x^{aug}(h,w) &= \mathcal{F}^{-1}(\mathcal{F}_s(x))(h,w)\\
    &=\frac{1}{HW}\sum_{u=0}^{H-1}\sum_{v=0}^{W-1}\mathcal{F}_s(x)(u,v) e^{-j2\pi (\frac{uh}{H}+\frac{vw}{W})}.
\end{aligned}
\label{eq:aug_x}
\end{equation}
The $\mathcal{F}(\cdot)$ and its inverse procedure $\mathcal{F}^{-1}(\cdot)$ can be efficiently implemented by the fast Fourier transform (FFT) and the inverse fast Fourier transform (IFFT) algorithms~\cite{frigo1998fftw}.
Finally, we randomly select one channel from the enhanced RGB image and duplicate it three times to suppress residual color bias after frequency exchange, while the enhanced IR image remains unmodified. This removes leftover RGB-specific chromatic cues so that the network focuses on the transferred illumination pattern and preserved structure. In this way, the illumination patterns of the enhanced RGB and IR images become similar, reducing the difficulty of subsequent cross-modal feature learning.

%

\subsection{Multi-Frequency Expert Network} 
\label{sec:mfen}
The RFA module initially reduces the discrepancy between RGB and IR data at the image level. 
We propose the Multi-Frequency Expert Network (MFEN) to capture frequency-domain information and further narrow the modality gap at the feature level.
MFEN constructs multiple experts for frequency-domain modulation, each handling a different frequency band. Their outputs are then integrated through a gating function.

\paragraph{Expert Design.}
Given a feature map $X\in \mathbb{R}^{H_X\times W_X \times C_X}$ as the input to each expert, two $1 \times 1$ convolutions project it to obtain the feature maps $Q$ and $K$, where $\{Q, K\}\in\mathbb{R}^{H_X\times W_X \times C'_X}$ and $C'_X$ is set to 64 so that the expert module remains lightweight and efficient.
$Q$ and $K$ are transformed by Fourier transform $\mathcal{F}(\cdot)$ on the spatial dimension to obtain their frequency domain representations:
\begin{equation}   
\begin{aligned} 
    Q_F=\mathcal{F}(Q),\quad K_F=\mathcal{F}(K).
\end{aligned}
\label{eq:qf,kf}
\end{equation}

To capture cues at distinct frequency bands, we employ a filter to eliminate non-target frequency components:
\begin{equation}   
\begin{aligned} 
    K_{Fb} = M_b\odot K_F,
\end{aligned}
\label{eq:sfa_hl}
\end{equation}
where $\odot$ denotes element-wise multiplication, $M_b \in \mathbb{R}^{H_X\times W_X}$ is a binary mask to isolate specific frequency ranges.
We apply the band mask only to $K_F$ while keeping $Q_F$ unchanged. In this way, $Q_F$ serves as a full-spectrum content anchor, and each expert learns how the complete feature content interacts with a target band. Masking both $Q_F$ and $K_F$ would restrict each expert to one narrow band and weaken cross-band complementarity.
We set $M_{b}(u,v)=1$ if $\frac{H_X}{2}\phi_b<|u-\frac{H_X}{2}|<\frac{H_X}{2}\phi_{b+1}$ and $\frac{W_X}{2}\phi_b<|v-\frac{W_X}{2}|<\frac{W_X}{2}\phi_{b+1}$, and assign zero to all other positions $(u,v)$.
Here, $\phi_b$ and $\phi_{b+1}$ are thresholds from the predefined frequency set $\{0,\phi_1,\phi_2,\ldots,1\}$.  
By default, we decompose the frequency spectrum into four distinct bands using an octave-based partitioning strategy~\cite{chen2025frequency}. 
For $n$ experts, the thresholds are $\{0,\frac{1}{2^{n-1}},\ldots,\frac{1}{4},\frac{1}{2},1\}$.
This octave-style design is suitable for VI-ReID because illumination-dominated low frequencies require coarse modeling, whereas identity-related details in higher frequencies benefit from finer partitioning. In practice, we use $n=4$ to balance these factors while keeping each expert expressive.
Subsequently, we modulate the target frequency components to extract frequency-domain cues, which are then transformed back to the spatial domain via the inverse transform:
\begin{equation}   
\begin{aligned}
    A_b = BN_b(\mathcal{F}^{-1}(Q_{F}\odot K_{Fb})),
\end{aligned}
\label{eq:sfa_ahl}
\end{equation}
where $BN_b(\cdot)$ is a batch normalization layer.

\paragraph{Frequency Fusion.}
To enable the model to adaptively capture different frequencies, we input the feature map $X$ into all experts, resulting in outputs $A_1, A_2, \ldots, A_n$.
Then we combine these experts through a gating network, producing the output $A$. More precisely, $A$ is the weighted sum of the outputs from the $n$ experts:
\begin{equation}
\label{eq:moe}
\begin{aligned}
    A = \sum\limits_{j=1}^n Gate(X)_j A_j,
\end{aligned}
\end{equation}
where $Gate(X) \in \mathbb{R}^n$ contains the weights of the $n$ experts and is calculated by a gating network:
\begin{equation}
\label{eq:gate}
\begin{aligned}
    Gate(X) = \mathrm{sigmoid}(W_g(X)),
\end{aligned}
\end{equation}
where $W_g(\cdot)$ is a 1 $\times$ 1 convolution for gate prediction.
Unlike popular mixture-of-experts methods, we do not select top-$k$ experts or normalize the weights $Gate(X)$.
This is because we aim for the model to learn from all frequency components simultaneously, and the non-overlapping bands make the experts complementary rather than competitive.
Under challenging lighting, multiple bands can be informative for the same sample, e.g., low frequencies for illumination correction and high frequencies for boundary recovery. Thus, top-$k$ selection would force unnecessary sparsity, while normalization would introduce competition among complementary experts.
Finally, we use the expert output $A$ to modulate the spatial-domain feature $X$:
\begin{equation}   
\begin{aligned} 
    X_{out} = W(A\odot W_V(X)),
\end{aligned}
\label{eq:sfa_end}
\end{equation}
where $W(\cdot)$ is a $1 \times 1$ convolution that maps the channels from $C'_X$ back to $C_X$, while $W_V(\cdot)$ is a $1 \times 1$ convolution that maps the channels from $C_X$ to $C_X'$.

In summary, we have devised a novel frequency modulation method, which can flexibly utilize binary frequency band masks for capturing specific frequency information. Additionally, the frequency fusion based on gating functions allows different samples to capture the most relevant frequency cues, providing greater flexibility compared to existing fixed-frequency methods.

\subsection{Frequency Auxiliary Optimization}
\label{sec:feo}
The feature map output by the model is $F\in \mathbb{R}^{H_F\times W_F\times C_F}$.
We obtain the feature $f=\mathrm{GAP}(F)$ through global average pooling (GAP).
Existing methods calculate the ReID loss with $f$ in the spatial domain to optimize the model.
We propose that introducing frequency-domain constraints during model optimization can further enhance cross-modal feature learning. 
Initially, we compute the frequency representation $f'$ of the features, which can be expressed as:
\begin{equation}   
\begin{aligned} 
    F' = \mathcal{F}(F),\quad f'=\mathrm{GAP}(F')
\end{aligned}
\label{eq:feo_f}
\end{equation}
To extract more comprehensive frequency information, we enhance the frequency representation by incorporating second-order moment information:
\begin{equation}   
\begin{aligned} 
    f''=f'+\sqrt{\mathrm{GAP}((F'-f')^2)}
\end{aligned}
\label{eq:feo_f2}
\end{equation}
The additional second-order term measures the dispersion of the frequency response and acts as an energy-aware complement to the first-order average $f'$. Therefore, $f''$ captures not only which frequency components are activated, but also how strongly they respond spatially.

The core idea of FAO is to impose loss constraints simultaneously on both $f$ and $f''$.
Importantly, FAO does not treat the frequency-domain representation as an isolated branch. Instead, $f''$ is a complementary statistical view that regularizes the main spatial feature $f$ during optimization.
We replace the conventional identity loss with the proposed \textbf{Frequency Identity Loss $\mathcal{L}_{fid}$}, which can be calculated as follows:
\begin{equation}   
\begin{aligned} 
    \mathcal{L}_{fid} &= \mathbb{E}_i[-y_i log (\frac{p_i+p''_i}{2})],
\end{aligned}
\label{eq:feo_fid}
\end{equation}
where $y_i$ is the identity label, and $p_i$ and $p''_i$ are the predicted probabilities for $f$ and $f''$ of the $i$-th image.
$p_i=\mathrm{softmax}(\mathrm{Cls}(f))$ and $p''_i=\mathrm{softmax}(\mathrm{Cls_2}(f''))$, where $\mathrm{Cls}(\cdot)$ and $\mathrm{Cls_2}(\cdot)$ are two classifiers, and $\mathrm{softmax}(\cdot)$ is the softmax function.

To further eliminate modality differences, we introduce the \textbf{Frequency KL Divergence Loss $\mathcal{L}_{fkl}$}, which can be calculated as follows:
\begin{equation}   
\begin{aligned} 
    \mathcal{L}_{fkl} & = \mathbb{E}_i[\mathrm{D_{kl}}(\frac{p_i+p''_i}{2}||\mathrm{AVG_+}(\frac{p_j+p''_j}{2}))],
\end{aligned}
\label{eq:feo_fkl}
\end{equation}
where $\mathrm{D_{kl}}(\cdot||\cdot)$ denotes the KL divergence, and $\mathrm{AVG_+}(\frac{p_j+p''_j}{2})$ is computed by averaging the classification probabilities from images with a different modality and the same identity as the $i$-th image in the training batch.
Intuitively, $\mathcal{L}_{fkl}$ encourages cross-modal positive samples to exhibit similar classification probabilities.
This is natural because $f$ and $f''$ describe the same identity from complementary spatial-frequency views.

Inspired by~\cite{wu2021discover,li2022counterfactual}, we propose the \textbf{Frequency Euclidean Distance Loss $\mathcal{L}_{feu}$} to impose frequency domain feature constraints within the Euclidean space, which can be calculated as follows:
\begin{equation}   
\begin{aligned} 
    \mathcal{L}_{feu} & = \mathbb{E}_i[\mathrm{D_{eu}}(\frac{f_i+f''_i}{2},\mathrm{AVG_+}(\frac{f_j+f''_j}{2}))]\\
    &+ \mathbb{E}_i[\text{max}(0,\rho - \mathrm{D_{eu}}(\frac{f_i+f''_i}{2},\mathrm{AVG_-}(\frac{f_k+f''_k}{2})))],
\end{aligned}
\label{eq:feo_feu}
\end{equation}
where $\mathrm{D_{eu}}(\cdot,\cdot)$ is the Euclidean distance, and $\rho$ is the margin parameter and is set to 0.6. 
$\mathrm{AVG_+}(\frac{f_j+f''_j}{2})$ is computed by averaging the features from images with a different modality and the same identity as the $i$-th image in the training batch.
Meanwhile, $\mathrm{AVG_-}(\frac{f_k+f''_k}{2})$ is computed by averaging the features from images with a different modality and a different identity from the $i$-th image in the training batch.
Intuitively, $\mathcal{L}_{feu}$ encourages positive samples across modalities to get closer to each other while negative samples are pushed farther.
Likewise, the Euclidean constraint refines the geometry of the final embedding with complementary frequency statistics rather than replacing spatial-domain learning.

The whole model is trained end-to-end and the total loss $\mathcal{L}_{total}$ of our method is defined as:
\begin{equation}
\begin{aligned}
\mathcal{L}_{total}=\mathcal{L}_{fid}+\mathcal{L}_{fkl}+\mathcal{L}_{feu}.
\label{eq:feo}
\end{aligned}
\end{equation}
Overall, FAO encourages the model to extract discriminative frequency-domain cues as a complement to spatial-domain features. During inference, we concatenate the spatial feature $f$ with its complementary frequency statistic $f''$ for retrieval.

\section{Experiments}
\label{sec:exper}

\begin{table*}[!t]
    \fontsize{9}{9}\selectfont
    \setlength\tabcolsep{7.5pt}
    \renewcommand{\arraystretch}{1.4}
    \centering
    \caption{Comparison of rank-k (R-k) accuracy (\%) and mAP accuracy (\%) with the state-of-the-art methods on the SYSU-MM01 and the RegDB datasets. The best results are marked in bold.
    }
    \vspace{0mm}
    \begin{tabular}{l c c c c c c c c c c c c}
        \toprule[1pt]
        &
        \multicolumn{6}{c}{\bf SYSU-MM01}&
        \multicolumn{6}{c}{\bf RegDB}
        \cr\cmidrule(r){2-13}&
        \multicolumn{3}{c}{\bf All-Search}&
        \multicolumn{3}{c}{\bf Indoor-Search}&
        \multicolumn{3}{c}{\bf Visible to Infrared}&
        \multicolumn{3}{c}{\bf Infrared to Visible}
        \cr\cmidrule(r){2-13}
        \multirow{-3}{*}{\bf Method}&
        \textbf{R-1}&\textbf{R-10}&\textbf{mAP}&
        \textbf{R-1}&\textbf{R-10}&\textbf{mAP}&
        \textbf{R-1}&\textbf{R-10}&\textbf{mAP}&
        \textbf{R-1}&\textbf{R-10}&\textbf{mAP}
        \cr\midrule
        AGW~\cite{ye2021deep}&
        47.50&84.39&47.65&54.17&91.14&62.97&
        70.05&-&66.37&70.49&-&65.90\cr
        DDAG~\cite{ye2020dynamic}&
        54.75&90.39&53.02&61.02&94.06&67.98&
        69.34&86.19&63.46&68.06&85.15&61.80\cr
        cm-SSFT~\cite{lu2020cross}&
        61.6&89.2&63.2&70.5&94.9&72.6&
        72.3&-&72.9&71.0&-&71.7\cr
        CAJ~\cite{ye2021channel}&
        69.88&95.71&66.89&76.26&97.88&80.37&
        85.03&95.49&79.14&84.75&95.33&77.82\cr
        MPANet~\cite{wu2021discover}&
        70.58&96.21&68.24&76.74&98.21&80.95&
        83.7 &-&80.9&82.8&-&80.7\cr
        MMN~\cite{zhang2021towards}&
        70.6&96.2&66.9&76.2&97.2&79.6&
        91.6&97.7&84.1&87.5&96.0&80.5\cr
        CIFT$^\dagger$~\cite{li2022counterfactual}&
        71.77&-&67.64&78.65&-&82.11&
        92.17&-&86.96&90.12&-&84.81\cr
        DEEN~\cite{zhang2023diverse}&
        74.7&97.6&71.8&80.3&99.0&83.3&
        91.1&97.8&85.1&89.5&96.8&83.4\cr
        PMCM~\cite{qian2025visible}&
        75.54&97.49&71.16&81.52&98.99&84.33&
        93.09&-&89.57&91.44&-&87.15\cr
        MSCLNet~\cite{zhang2022modality}&
        76.99&97.63&71.64&78.49&99.32&81.17&
        84.17&-&80.99&83.86&-&78.31\cr
        SGIEL~\cite{fang2023visible}&
        77.12&97.03&72.33&82.07&97.42&82.95&
        92.18&-&86.59&91.07&-&85.23\cr
        PartMix~\cite{kim2023partmix}&
        77.78&-&74.62&81.52&-&84.38&
        85.66&-&82.27&84.93&-&82.52\cr
        DNS~\cite{jiang2024domain}&
        77.27&-&74.35&84.21&-&86.83&
        93.01&-&88.56&93.48&-&88.10\cr
        ProtoHPE~\cite{zhang2023protohpe}&
        71.92&96.19&70.59&77.81&98.64&81.31&         
        88.74&-&83.72&88.69&-&81.99\cr
        FDMNet~\cite{li2024frequency}&
        75.99&97.63&70.71&80.92&98.88&82.64&
        \textbf{95.92}&99.01&89.26&93.58&98.33&86.88\cr
        FDNM~\cite{zhang2024frequency}&
        77.8&97.8&75.1&87.3&99.4&\textbf{89.1}&
        95.5&99.0&90.0&94.0&98.5&88.7\cr
        DSSF$^3$~\cite{chen2024visible}&
        79.12&\textbf{97.98}&75.27&85.01&99.17&86.75&
        91.91&97.40&86.24&90.45&97.07&85.12\cr
        \midrule
        MFEN (Ours)&
        \textbf{80.93}&97.12&
        \textbf{76.56}&\textbf{87.88}&
        \textbf{99.43}&88.12&
        94.85&\textbf{99.06}&
        \textbf{90.06}&\textbf{94.11}&
        \textbf{98.83}&\textbf{90.25}\cr		
        \bottomrule[1pt]
    \end{tabular}
    \vspace{0mm}
    \label{tab:comp}
\end{table*}

\paragraph{Datasets.}
We conduct experiments on three challenging VI-ReID datasets: SYSU-MM01~\cite{wu2017rgb}, RegDB~\cite{nguyen2017person}, and LLCM~\cite{zhang2023diverse}. SYSU-MM01 contains 395/96 train/test identities and is evaluated under all-search and indoor-search settings~\cite{ye2021deep}. RegDB contains 412 identities split into 206/206 train/test identities and is evaluated in visible-to-infrared and infrared-to-visible modes~\cite{ye2021deep}. LLCM contains 713/351 train/test identities and is also evaluated in the two retrieval directions~\cite{zhang2023diverse}.

\paragraph{Evaluation Protocol.}
We follow the standard evaluation protocols in existing VI-ReID benchmarks~\cite{ye2021deep}. Results on SYSU-MM01 and LLCM are averaged over 10 random gallery splits, while those on RegDB are averaged over 10 trials with different training/testing splits. We report cumulative matching characteristics (CMC) and mean average precision (mAP).

\paragraph{Implementation Details.}
We use a ResNet-50~\cite{he2016deep} backbone pre-trained on ImageNet.
Following~\cite{luo2019bag}, we set the last convolutional stride to 1 and add BNNeck.
MFEN is inserted after the second and third layers of the network.
Following~\cite{chen2024visible,zhang2024frequency,zhang2023mrcn}, each image is resized to 384$\times$192 and augmented with random cropping, random horizontal flipping, and random erasing~\cite{zhong2020random}.
We train the whole model with SGD for 120 epochs using a batch size of 64 with 8 identities, a learning rate of 0.02, warm-up, and cosine decay.
In all experiments, MFEN uses four experts and the margin parameter is set to $\rho=0.6$.

\begin{table}[!h]
    \fontsize{9}{9}\selectfont
    \setlength\tabcolsep{5.0pt}
    \renewcommand{\arraystretch}{1.4}
    \centering
    \caption{Comparison of rank-k (R-k) accuracy (\%) and mAP accuracy (\%) with the state-of-the-art methods on the LLCM dataset. The best results are marked in bold.
    }
    \vspace{0mm}
    \begin{tabular}{l c c c c c c}
        \toprule[1pt]
        &
        \multicolumn{6}{c}{\bf LLCM}
        \cr\cmidrule(r){2-7}&
        \multicolumn{3}{c}{\bf Visible to Infrared}&
        \multicolumn{3}{c}{\bf Infrared to Visible}
        \cr\cmidrule(r){2-7}
        \multirow{-3}{*}{\bf Method}&
        \textbf{R-1}&\textbf{R-10}&\textbf{mAP}&
        \textbf{R-1}&\textbf{R-10}&\textbf{mAP}
        \cr\midrule
        DDAG~\cite{ye2020dynamic}&         48.0&79.2&52.3&40.3&71.4&48.4\cr
        AGW~\cite{ye2021deep}&
        51.5&81.5&55.3&43.6&74.6&51.8\cr
        CAJ~\cite{ye2021channel}&
        56.5&85.3&59.8&48.8&79.5&56.6\cr
        MMN~\cite{zhang2021towards}&
        59.9&88.5&62.7&52.5&81.6&58.9\cr
        DEEN~\cite{zhang2023diverse}&
        62.5&90.3&65.8&54.9&84.9&62.9\cr
        DNS~\cite{jiang2024domain}&         
        66.0&-&68.6&57.5&-&64.1\cr
        \midrule
        MFEN (Ours)&
        \textbf{67.9}&\textbf{91.0}&\textbf{69.8}&
        \textbf{59.0}&\textbf{85.6}&\textbf{65.3}\cr		
        \bottomrule[1pt]
    \end{tabular}
    \vspace{0mm}
    \label{tab:comp2}
\end{table}

\subsection{Comparison with SOTA methods}

In this part, we compare MFEN with state-of-the-art (SOTA) VI-ReID approaches.
%
The comparison results on SYSU-MM01 and RegDB are shown in \cref{tab:comp}. MFEN outperforms existing SOTA methods or achieves comparable performance across all settings. On SYSU-MM01, MFEN achieves 80.93\% rank-1 accuracy and 76.56\% mAP accuracy, surpassing DSSF$^3$~\cite{chen2024visible} by 1.81\% and 1.29\% in the most challenging all-search mode. Since all-search contains both indoor and outdoor scenes with more diverse illumination and clutter, this larger gain supports our motivation for multi-band modeling and adaptive sample-wise fusion.
On RegDB, the average performances of MFEN in the `Visible to Infrared' and `Infrared to Visible' modes are 94.48\% rank-1 accuracy and 90.16\% mAP accuracy, surpassing DSSF$^3$~\cite{chen2024visible} by about 3.30\% on rank-1 accuracy and 4.48\% on mAP accuracy. These gains across both retrieval directions indicate that our method is not a dataset-specific trick on SYSU-MM01.
The comparison result on LLCM is shown in \cref{tab:comp2}. The average performances of MFEN in the two testing modes are 63.5\% rank-1 accuracy and 67.6\% mAP accuracy, surpassing DNS~\cite{jiang2024domain} by about 1.7\% on rank-1 accuracy and 1.2\% on mAP accuracy. Since LLCM contains more challenging night-time scenes, this result further shows that the proposed frequency modeling remains stable under highly complex lighting.
Compared with prior frequency-based methods~\cite{zhang2023protohpe,li2024frequency,zhang2024frequency,chen2024visible}, MFEN improves performance by combining multi-band mining with sample-wise fusion instead of relying on full-spectrum modulation or a fixed band.

\begin{table}[t]
    \fontsize{9}{9}\selectfont     
    \setlength\tabcolsep{8.5pt}
    \renewcommand{\arraystretch}{1.4}
    \centering
    \caption{Ablation study about each component of our MFEN on the SYSU-MM01 dataset all-Search mode.
    The rank-k (R-k) accuracy (\%) and mAP accuracy (\%) are reported.
    }
    \vspace{0mm}
    \begin{tabular}{c ccc cc}
        \toprule[1pt]
        {\bf i}
        &\textbf{RFA}&\textbf{MFEN}&
        \textbf{FAO}&\textbf{R-1}&\textbf{mAP}
        \cr\midrule
        1&&&&
        71.85&68.95
        \cr
        2&\Checkmark&&&
        75.01&71.23
        \cr
        3&\Checkmark&\Checkmark&&         
        78.42&74.85         
        \cr
        4&\Checkmark&\Checkmark&\Checkmark& 
        80.93&76.56
        \cr
        \bottomrule[1pt]
    \end{tabular}
    \vspace{0mm}
    \label{tab:abl}
\end{table}

\begin{table}[t]
    \fontsize{9}{9}\selectfont     
    \setlength\tabcolsep{8.0pt}
    \renewcommand{\arraystretch}{1.4}
    \centering
    \caption{Ablation study about our RFA on the SYSU-MM01 dataset all-Search mode.
    The rank-k (R-k) accuracy (\%) and mAP accuracy (\%) are reported.
    }     
    \vspace{0mm}
    \begin{tabular}{c l cc}
        \toprule[1pt]
        {\bf i}
        &\textbf{RFA}&\textbf{R-1}&\textbf{mAP}
        \cr\midrule
        1&RFA$\rightarrow$No Aug&
        77.41&72.99\cr
        2&RFA$\rightarrow$Random Gray+Brightness&         78.85&74.06\cr
        3&RFA$\rightarrow$CAJ~\cite{ye2021channel}&
        79.88&75.85\cr
        4&RFA$\rightarrow$DMT~\cite{cui2024dma}&         79.19&74.70\cr
        5&RFA$\rightarrow$RLE~\cite{tan2024rle}&
        79.87&75.48\cr
        \midrule
        6&RFA (Ours)&
        80.93&76.56\cr
        \bottomrule[1pt]
    \end{tabular}
    \vspace{0mm}
    \label{tab:fda}
\end{table}

\subsection{Ablation Study}

In this section, we conduct ablation studies to evaluate the effectiveness of each module, i.e. RFA, MFEN, and FAO.
The baseline method uses ResNet-50 as the backbone network and is trained with the identity loss, the KL divergence loss, and the Euclidean distance loss.
Specifically, the three losses of this baseline correspond to the three losses in our FAO, but excluding all frequency-related components.
All ablation experiments are performed on our baseline in the all-search mode of the SYSU-MM01 dataset.

\paragraph{Effectiveness of the Overall Method.}
To illustrate the contribution of each module, we add them into the model one by one.
The results are shown in \cref{tab:abl}.
Compared with the baseline, the RFA improves the rank-1 and mAP accuracy by 3.16\% and 2.28\%, verifying the benefit of reducing image-level illumination discrepancy before feature extraction.
MFEN further improves rank-1 and mAP by 3.41\% and 3.62\%, showing that adaptive multi-band fusion is the core contributor.
Finally, FAO improves rank-1 and mAP by another 2.51\% and 1.71\%, indicating that frequency-aware optimization further refines the learned representation. These results show that our approach incorporates frequency information effectively at the data, model, and optimization levels.

\begin{table}[t]
    \fontsize{9}{9}\selectfont     
    \setlength\tabcolsep{9.0pt}
    \renewcommand{\arraystretch}{1.4}
    \centering
    \caption{Ablation study about our MFEN on the SYSU-MM01 dataset all-Search mode.
    The rank-k (R-k) accuracy (\%) and mAP accuracy (\%) are reported.
    }
    \vspace{0mm}
    \begin{tabular}{c l cc}
        \toprule[1pt]
        {\bf i}
       &\textbf{MFEN}&\textbf{R-1}&\textbf{mAP}
        \cr\midrule
        1&MFEN$\rightarrow$SE~\cite{hu2018squeeze}&
        76.90&71.55\cr
        2&MFEN$\rightarrow$CBAM~\cite{woo2018cbam}&            75.45&70.33\cr
        3&MFEN$\rightarrow$Non-Local~\cite{wang2018non,ye2021deep}&
        78.96&75.80\cr\midrule
        4&1 Expert&                                    79.87&75.61\cr
        5&2 Experts&                                    80.22&75.93\cr
        6&1 High-frequency Expert&
        79.63&74.92\cr
        7&1 Low-frequency Expert& 
        78.52&74.32\cr\midrule
        8&MFEN After Layer 1&                         78.66&74.70\cr         
        9&MFEN After Layer 2&                         79.95&76.00\cr         
        10&MFEN After Layer 3&         
        80.11&76.07\cr         
        11&MFEN After Layer 4&          
        76.42&72.05\cr\midrule
        12&MFEN (Ours)&                           
        80.93&76.56\cr
        \bottomrule[1pt]
    \end{tabular}
    \vspace{0mm}
    \label{tab:cfa}
\end{table}

\paragraph{Effectiveness of RFA.}
As shown in \cref{tab:fda}, we replace our RFA with other data augmentation methods to individually analyze the effect of the RFA module.
The results in the first row show that omitting data augmentation leads to a decrease in rank-1 accuracy by 3.52\% and a reduction in mAP by 3.57\%.
The second row demonstrates that randomly converting images to grayscale and randomly changing image brightness by [-40\%, +40\%] is less effective than our RFA.
This is because manually setting color and brightness variations struggles to accurately simulate real-world lighting conditions. In contrast, our RFA effectively mitigates lighting differences by exchanging the amplitude of image frequency domain representations.

Additionally, RFA outperforms the widely adopted channel augmentation method, CAJ~\cite{ye2021channel}, and other augmentation methods~\cite{tan2024rle,cui2024dma}.
These methods primarily focus on color changes and the overall brightness variations of images. In contrast, our approach can generate local brightness variations that effectively simulate lighting conditions such as overexposure and underexposure.

\begin{table}[t]  
    \fontsize{9}{9}\selectfont     
    \setlength\tabcolsep{9.0pt}
    \renewcommand{\arraystretch}{1.4}     
    \centering
    \caption{Ablation study about our FAO on the SYSU-MM01 dataset all-Search mode.
    The rank-k (R-k) accuracy (\%) and mAP accuracy (\%) are reported.
    }     
    \vspace{0mm}
    \begin{tabular}{c l cc}
        \toprule[1pt]
        {\bf i}
        &\textbf{FAO}&\textbf{R-1}&\textbf{mAP}
        \cr\midrule
        1&$\mathcal{L}_{id}$&67.80&66.41\cr
        
        2&$\mathcal{L}_{fid}$&69.02&67.55\cr\midrule
        
        3&$\mathcal{L}_{fid}$ + $\mathcal{L}_{kl}$&
        76.23&72.81\cr
        
        4&$\mathcal{L}_{fid}$ + $\mathcal{L}_{fkl}$&
        78.05&74.91\cr\midrule
        
        5&$\mathcal{L}_{fid}$ + $\mathcal{L}_{fkl}$ + $\mathcal{L}_{eu}$&
        79.72&75.02\cr         
        
        6&$\mathcal{L}_{fid}$ + $\mathcal{L}_{fkl}$ + $\mathcal{L}_{feu}$ (Ours)&
        80.93&76.56\cr
        \bottomrule[1pt]
    \end{tabular}
    \vspace{0mm}
    \label{tab:feo}
\end{table}

\paragraph{Effectiveness of MFEN.}
%
To demonstrate the advantages of MFEN, we compare it with existing attention methods, including the SE block~\cite{hu2018squeeze}, CBAM~\cite{woo2018cbam}, and the Non-local block~\cite{wang2018non} (following AGW~\cite{ye2021deep}). 
%
%

As shown in \cref{tab:cfa}, replacing MFEN with the SE block results in a decline in rank-1 and mAP accuracy by 4.03\% and 5.01\%, respectively.
Compared with MFEN, CBAM reduces rank-1 and mAP accuracy by 5.48\% and 6.23\%, respectively.
The MFEN performs element-wise multiplication in the frequency domain and then transforms the result back to the spatial domain.
This process allows for the interaction of features from different locations, which is a capability not achieved by other element-wise multiplication methods such as the CBAM.
Replacing MFEN with the Non-local block results in a decline in rank-1 and mAP accuracy by 1.97\% and 0.76\%, respectively.
Although the performance gap is not very significant, the MFEN employs element-wise multiplication, which is much more efficient compared to the quadratic complexity of non-local blocks.


The results presented in lines 4 to 7 of \cref{tab:cfa} evaluate the setting of experts.
Experimental results from lines 4 and 5 indicate that utilizing either a full-spectrum expert or simply categorizing frequencies into two experts (high-frequency and low-frequency) results in suboptimal performance.
Additionally, the results from lines 6 and 7 demonstrate that focusing solely on either high-frequency or low-frequency may overlook critical information, resulting in poorer outcomes compared to utilizing a full frequency spectrum expert.
Together, these results validate our hypothesis that identity cues and illumination disturbances are distributed across multiple bands and that the useful band is not fixed across samples, explaining why our method outperforms existing frequency-based approaches~\cite{zhang2023protohpe,li2024frequency,zhang2024frequency,chen2024visible}.

The results in lines 8 to 11 of \cref{tab:cfa} evaluate the optimal integration locations of MFEN. They indicate that incorporating MFEN after the second and third layers of ResNet-50 yields the best improvements, suggesting that intermediate-layer features are more suitable for extracting discriminative frequency information.

\paragraph{Effectiveness of FAO.}
As shown in \cref{tab:feo}, we analyze the effects of the three losses in FAO: $\mathcal{L}_{fid}$, $\mathcal{L}_{fkl}$, and $\mathcal{L}_{feu}$.
To comprehensively investigate the role of frequency enhancement, we removed the frequency components (i.e., $f''$) from these three losses and denoted them as $\mathcal{L}_{id}$, $\mathcal{L}_{kl}$, and $\mathcal{L}_{eu}$.
Experimental results demonstrate that frequency domain enhancement consistently improves performance across all three types of losses.
This demonstrates that frequency enhancement has broad applicability and complements the spatial-domain loss; FAO is not tied to one specific supervision form, and the frequency statistics provide a useful auxiliary constraint for improving the final embedding geometry.

\subsection{Modality-wise Frequency Statistics}

To better illustrate our motivation, we quantify the image-level frequency amplitude distribution of RGB and IR images on SYSU-MM01.
For each image, we resize it to 384$\times$192, compute its FFT amplitude spectrum, and normalize the band-wise amplitude ratios.
Following the camera layout of SYSU-MM01, we separate the training images into four scene-modality groups: indoor RGB, outdoor RGB, indoor IR, and outdoor IR.
As shown in \cref{fig:intro,tab:freq_stat}, these groups exhibit clearly different band-wise amplitude distributions.
In particular, IR images concentrate more amplitude in the lowest-frequency range, while RGB images retain relatively richer amplitude in the mid- and high-frequency ranges.
The mid-to-high/low amplitude ratio further decreases from outdoor RGB (1.96) and indoor RGB (1.51) to indoor IR (1.18) and outdoor IR (1.01).
This observation is consistent with our motivation that the modality gap is closely related to illumination-induced frequency differences, and further justifies sample-wise multi-band modeling over uniform whole-spectrum processing.

\begin{table}[!t]
    \fontsize{9}{9}\selectfont
    \setlength\tabcolsep{4pt}
    \renewcommand{\arraystretch}{1.4}
    \centering
    \caption{Band-wise frequency amplitude distribution (\%) of different scene-modality groups on SYSU-MM01.}
    \vspace{0mm}
    \begin{tabular}{lcccc}
        \toprule[1pt]
        \textbf{Frequency} & \makecell{\textbf{Indoor}\\\textbf{RGB}} & \makecell{\textbf{Outdoor}\\\textbf{RGB}} & \makecell{\textbf{Indoor}\\\textbf{IR}} & \makecell{\textbf{Outdoor}\\\textbf{IR}} \\
        \midrule
        $0$--$1/16$ & 33.01 & 27.81 & 38.25 & 41.25 \\
        $1/16$--$1/8$ & 17.28 & 17.72 & 16.55 & 17.04 \\
        $1/8$--$1/4$ & 19.50 & 22.13 & 17.81 & 16.47 \\
        $1/4$--$1/2$ & 15.96 & 18.08 & 15.05 & 12.82 \\
        $1/2$--$1$ & 14.25 & 14.27 & 12.33 & 12.41 \\
        \bottomrule[1pt]
    \end{tabular}
    \vspace{0mm}
    \label{tab:freq_stat}
\end{table}

\section{Visualization}
\label{sec:visualization}

We provide qualitative visualizations to further interpret the proposed modules.
These results show how RFA changes image-level illumination, how MFEN assigns different frequency experts to complementary identity cues, and how FAO improves the feature geometry.

\subsection{Visualization of Random Frequency Augmentation}

\begin{figure}[!h]
    \centering
    \includegraphics[width=0.9\linewidth]{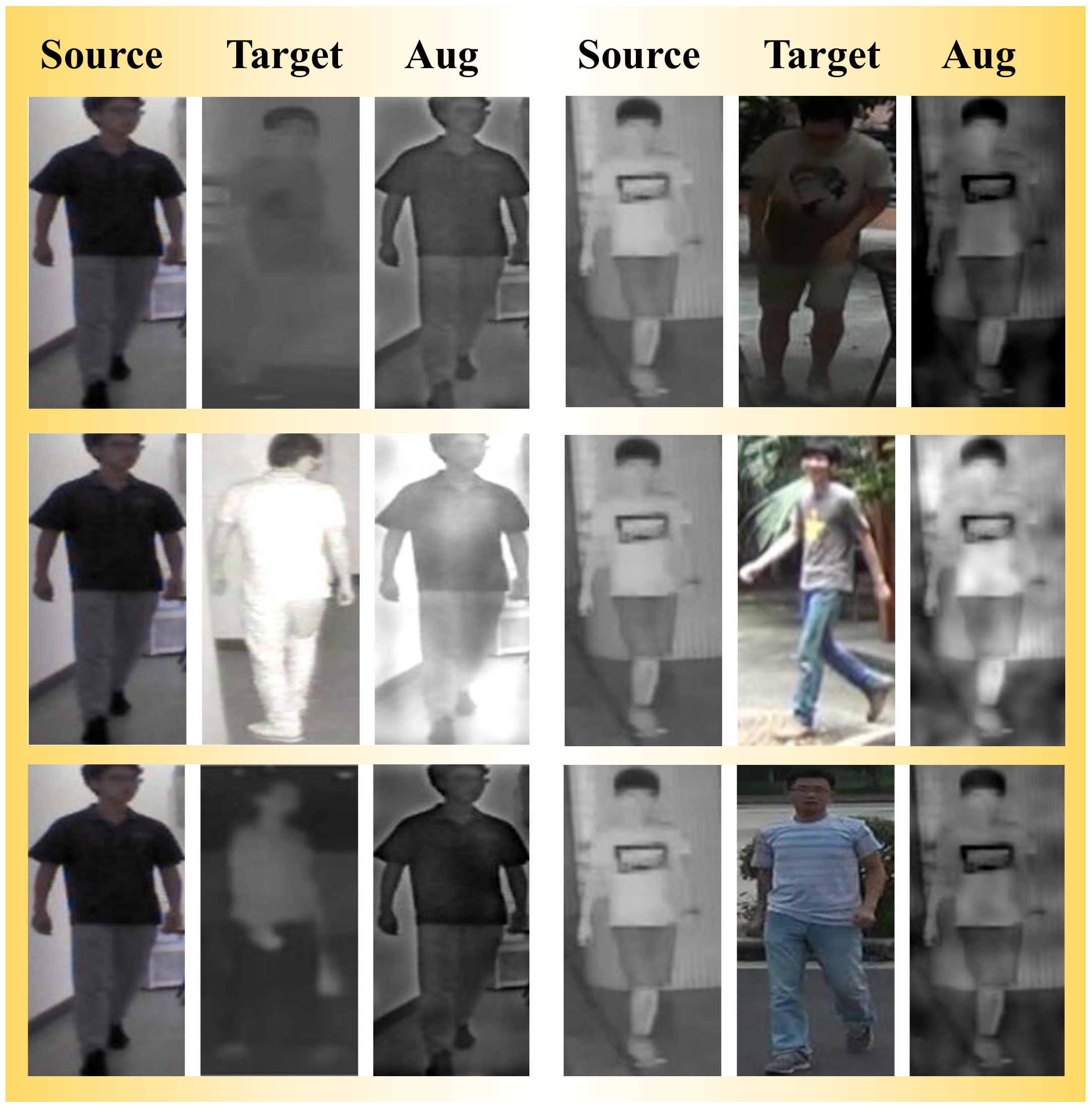}
    \caption{Visualization of Random Frequency Augmentation. The enhanced image (Aug) is obtained by exchanging the low-frequency amplitudes of the source image with the target image.}
    \label{fig:aug}
    \vspace{0mm}
\end{figure}

\cref{fig:aug} presents the visualization results of the proposed RFA.
The enhanced image (Aug) is obtained by exchanging the low-frequency amplitudes of the source image with the target image.
It is evident that, following RFA, the illumination conditions of the images become more diverse, effectively narrowing the gap between the two modalities.
At the same time, the main body structures of pedestrians remain recognizable, consistent with our design of exchanging only low-frequency amplitude while preserving structural details.

\subsection{Visualization of Multi-Frequency Expert Network}

\begin{figure}[!h]
    \centering
    \includegraphics[width=0.9\linewidth]{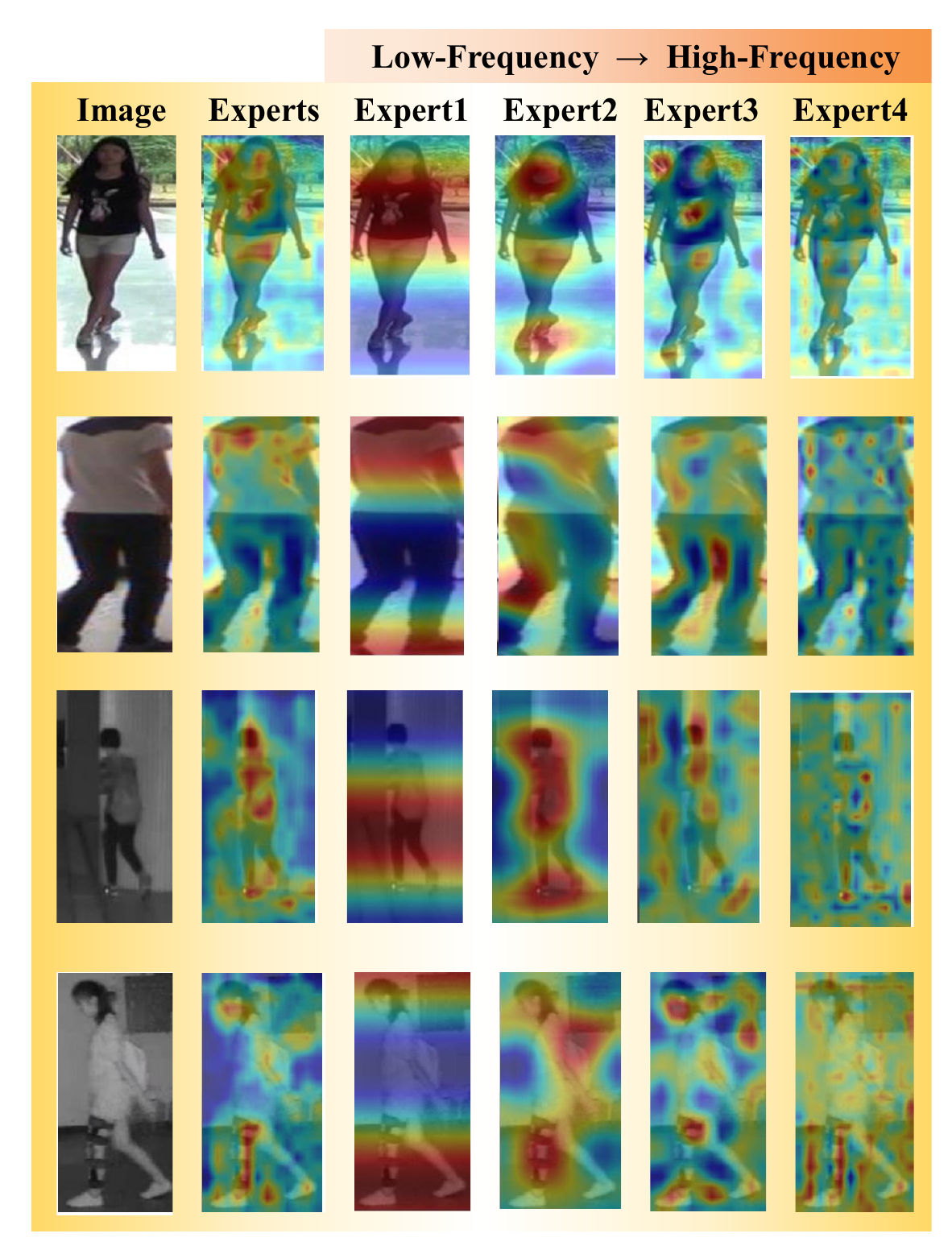}
    \caption{Visualization of the activation feature maps on the SYSU-MM01 dataset. In each row, the first column shows the original image, the second column shows the result obtained by combining all experts, while the remaining four columns correspond to the outputs of the four experts, arranged from low to high frequencies.}
    \label{fig:cam}
    \vspace{0mm}
\end{figure}

\cref{fig:cam} presents the activation feature maps of the proposed MFEN.
In each row, the first column shows the original image and the second column shows the result obtained by combining all experts, while the remaining four columns correspond to the outputs of the four experts, arranged from low to high frequencies.
Different experts, focusing on distinct frequency bands, capture identity cues of varying granularity.
Low-frequency experts tend to emphasize overall characteristics, whereas high-frequency experts focus on subtle variations.
Finally, when we dynamically combine all experts using a gating mechanism, the model focuses on the most discriminative regions.
This visualization directly supports the claim that the experts are not redundant: they learn complementary cues at different frequency bands, and the fusion module selects a more informative combination for each sample.

\subsection{Visualization of Frequency Auxiliary Optimization}

\begin{figure}[!h]
    \centering
    \includegraphics[width=1.0\linewidth]{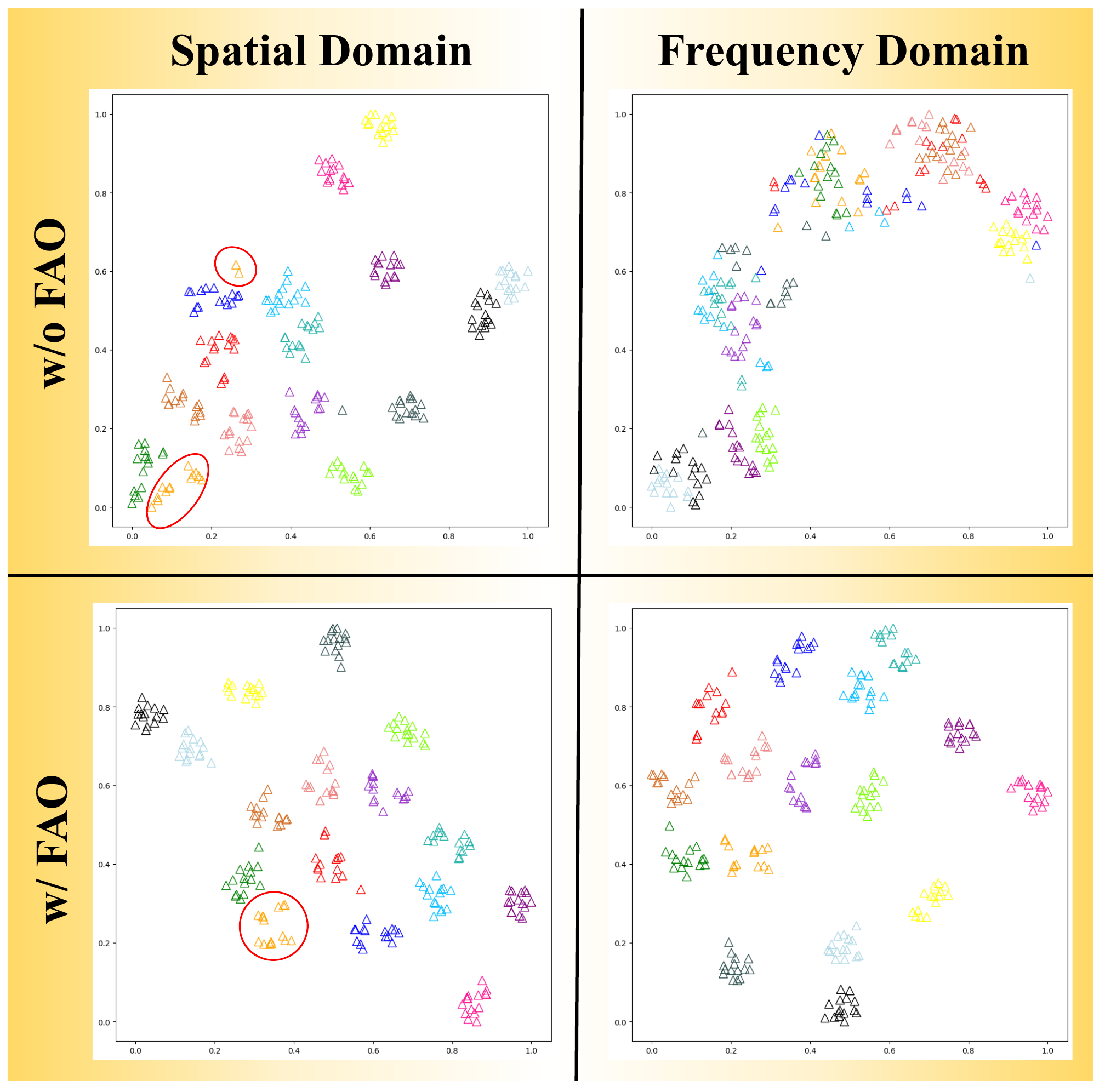}
    \caption{t-SNE~\citep{van2008visualizing} visualization of the distributions of image features from the spatial domain and the frequency domain. Different colors represent different identities.}
    \label{fig:tsne}
    \vspace{0mm}
\end{figure}

\cref{fig:tsne} presents the t-SNE~\citep{van2008visualizing} visualization of the distributions of image features from the spatial domain and the frequency domain.
In the absence of the proposed FAO, the spatial-domain features exhibit a certain degree of mis-clustering, while the frequency-domain features remain entirely indistinguishable.
However, when FAO is introduced, the additional frequency-domain constraints not only render the frequency distribution distinguishable but also collaboratively optimize the spatial feature distribution, resulting in more robust overall representations.
This observation is consistent with our formulation of FAO: frequency-domain statistics act as an auxiliary constraint that improves the geometry of the final embedding, rather than serving as an isolated branch.

\setcounter{section}{5}
\section{Conclusion}
\label{sec:conclusion}

In this paper, we propose a Multi-Frequency Expert Network (MFEN) for visible-infrared person re-identification (VI-ReID).
Our central insight is that illumination discrepancies in VI-ReID are not confined to a single fixed band, making multi-band modeling and sample-wise frequency fusion necessary. Accordingly, MFEN adaptively captures complementary cues from different bands, while RFA and FAO improve training from the data and optimization levels. Experiments on three challenging VI-ReID datasets demonstrate the effectiveness of the framework.

\section*{Acknowledgements}
This work is supported by the National Natural Science Foundation of China (Grant No. 62272430).

{
    \small
    \setlength{\bibsep}{0pt plus 0.2ex}
    \bibliographystyle{ieeenat_fullname}
    \bibliography{main}
}

\end{document}